\documentclass[conference]{IEEEtran}
\IEEEoverridecommandlockouts
\usepackage{cite}
\usepackage{amsmath,amssymb,amsfonts}
\usepackage{algorithmic}
\usepackage{graphicx}
\usepackage{textcomp}
\usepackage{xcolor}
\usepackage[hidelinks]{hyperref}
\usepackage{hypcap}
\def\BibTeX{{\rm B\kern .05em{\sc i\kern .025em b}\kern .08em
    T\kern .1667em\lower.7ex\hbox{E}\kern .125emX}}
\begin{document}

\title{A Masked Autoencoder Approach to Unsupervised Steel Surface Defect Recognition\\}

\author{
\IEEEauthorblockN{ Shrey Patel}
\IEEEauthorblockA{
\textit{University of Maryland, College Park}\\
United States of America \\
ORCID: 0009-0003-0886-4583}}

\maketitle

\begin{abstract}
Automated visual inspection of steel surface defects is a recurring quality control task in which labeled defect data is scarce and costly to obtain, while unlabeled surface images are abundant, which motivates self supervised methods that learn useful representations without class labels. A Transformer based Masked Autoencoder is used here to learn representations of steel surface defects for unsupervised grouping. During pretraining, 75\% of the input image patches are randomly masked, and a lightweight decoder reconstructs the masked regions from the visible 25\%. The encoder is trained jointly with an auxiliary defect localization objective, used only as a training signal and not evaluated as a detector. The decoder reaches a structural similarity score of 0.92 and a mean squared error of 0.47. Features from the pretrained encoder are then clustered using UMAP for dimensionality reduction and Agglomerative clustering, reaching a Hungarian matched accuracy of 91.3\% against the six known defect categories.
\end{abstract}


\section{Introduction}
Deep learning has become the dominant paradigm for defect recognition, with convolutional and, more recently, Transformer based architectures achieving strong results on benchmarks such as the NEU surface defect dataset across classification and detection formulations~\cite{b12}. However, these methods depend on annotations that are expensive to produce and must be repeated as products and defect distributions change, while unlabeled surface images are abundant. Self supervised learning addresses this by defining a pretext task whose signal comes from the data itself. This Masked Autoencoder(MAE)~\cite{b4}\cite{b18} approach masks a large fraction of image patches and reconstructs them from the visible remainder using an asymmetric encoder and decoder, yielding transferable representations well suited to the label scarce industrial setting~\cite{b5}.

This approach utilizes a Transformer based Masked Autoencoder which was pretrained on the NEU surface defect dataset and used the resulting encoder to group defect images into the six known categories without label supervision. Grayscale images are preprocessed with Contrast Limited Adaptive Histogram Equalization(CLAHE) and divided into fixed size patches for masked reconstruction pretraining. Features from the pretrained encoder are then passed to UMAP~\cite{b6} for dimensionality reduction, which later are grouped with Agglomerative clustering.

\section{Literature Review}
The NEU surface defect dataset introduced by Song and Yan \cite{b1}, comprising six categories of hot rolled steel defects, established an early foundation for learning based approaches, with a related benchmark and detection network later contributed by Lv \textit{et al.} \cite{b2}. Early automated systems relied on hand crafted texture descriptors such as local binary patterns and gray level features, but these proved brittle under the morphological and textural complexity of real defects, and deep learning displaced them as the dominant approach.

The dominant line of subsequent work has framed defect recognition as a supervised object detection or classification problem. Deep convolutional and, more recently, transformer based models have been applied to the NEU benchmark and its relatives \cite{b12}. A representative recent detector is MESC DETR \cite{b3}, which builds on the real time RT DETR framework and introduces a composite ConvNeXtV2 backbone, an edge enhanced feature fusion module targeting the blurred boundaries and extreme scale variation characteristic of metal defects, and a Focal MPDIoU regression loss, reporting improved mean average precision over the RT DETR baseline on both GC10 DET and NEU DET datasets while using Grad CAM++ to argue that its attention concentrates on defect regions. Transformer architectures have likewise been explored through Swin Transformer based designs for steel defect detection and classification \cite{b9}, while earlier work advanced supervised detection by fusing multiple hierarchical feature levels \cite{b11}. Across these approaches, performance rests on large quantities of manually annotated bounding boxes or class labels, costly to obtain in production settings where unlabeled surface images are far more plentiful.

A smaller body of work instead reduces dependence on annotation by learning directly from unlabeled defect imagery, including self supervised strategies investigated for surface defect tasks \cite{b10}. Among self supervised objectives, masked image modeling has proven especially effective for the Vision Transformer \cite{b5}, whose global self attention removes the locality constraints of convolutional architectures but which typically requires large datasets to train from scratch. The Masked Autoencoder \cite{b4} addresses this by masking a large fraction of image patches, encoding only the small visible subset, and reconstructing the missing content with a lightweight decoder. The resulting representations transfer well to downstream tasks, which makes the approach a natural candidate for the label scarce industrial setting.

Interpreting what such representations attend to is an established practice. Grad CAM \cite{b7}\cite{b17} and its generalization Grad CAM++ \cite{b8}\cite{b17} produce gradient based attention maps from a trained network's intermediate activations and have been applied to defect models, including on the NEU benchmark \cite{b3}, to assess whether learned attention aligns with visually salient defect regions. In the present work, features from a masked autoencoder pretrained encoder are reduced in dimension with UMAP \cite{b6} and grouped without label supervision, and encoder attention is examined using Grad CAM and Grad CAM++ to evaluate whether the learned representation localizes defects without explicit spatial supervision.

\section{Proposed Methodology}
The pretraining for the proposed model, a Vision Transformer based Masked Autoencoder, ran for approximately 2 hours and 20 minutes on Google Colab T4 GPU. Code availability: "https://anonymous.4open.science/status/NEU-Steel-Surface-Defect-315F". The NEU DET dataset is publicly available on Kaggle.

\subsection{Data Collection and Augmentation}\label{AA}
\begin{figure}[htbp]
\centerline{\includegraphics[width=0.5\textwidth]{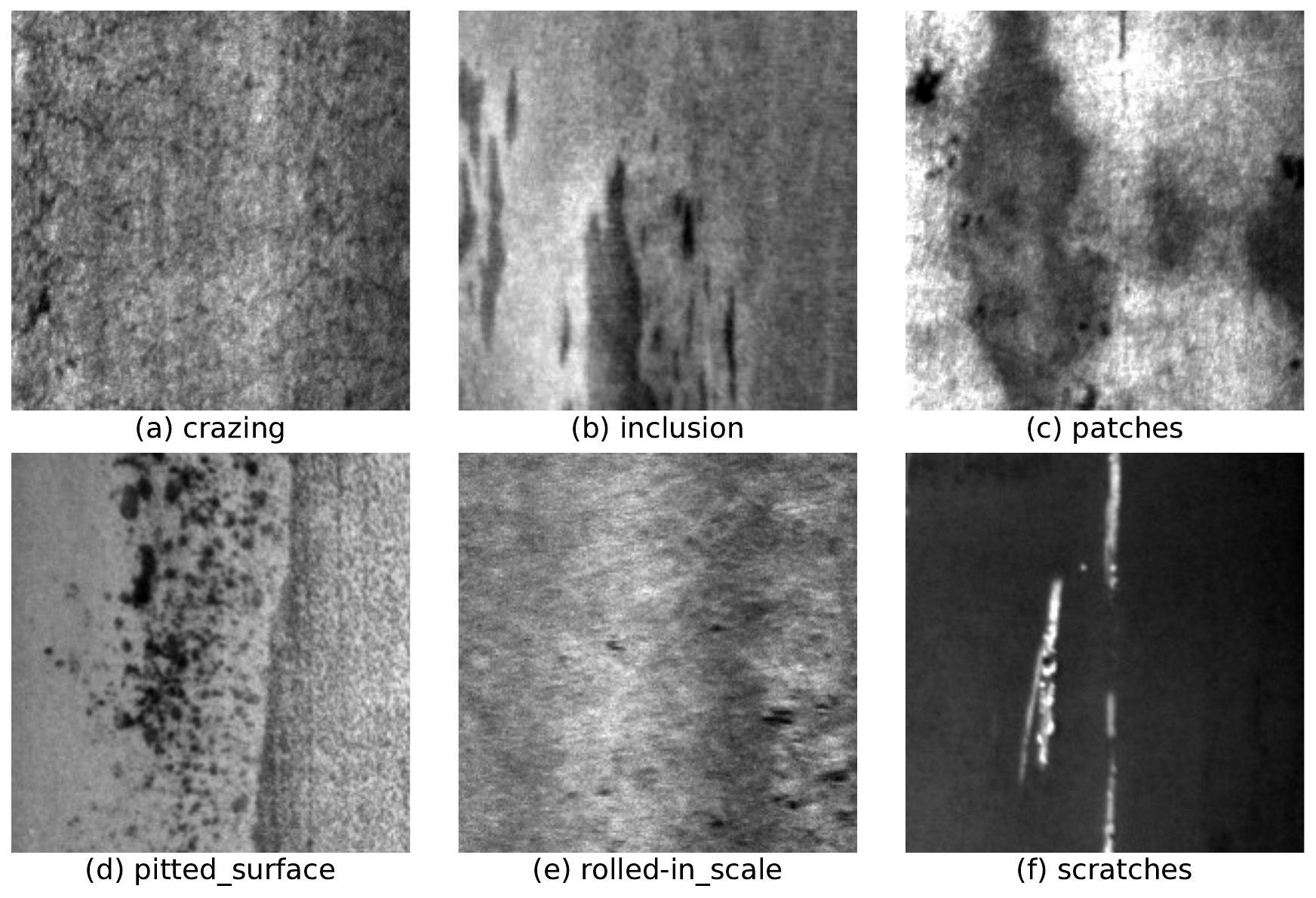}}
\caption{Sample of each labelled image present in the NEU-DET dataset}
\label{fig:dataset}
\end{figure}

The NEU DET dataset is used, which contains 1,800 grayscale hot rolled steel surface images evenly distributed across six defect categories: crazing, inclusion, patches, pitted surface, rolled in scale, and scratches. Representative samples of each category are shown in Figure~\ref{fig:dataset}.

Each image is loaded as grayscale and enhanced using CLAHE algorithm to improve local contrast, then resized to 224×224 pixels and normalized to the [0, 1] range. During training, images are augmented with transformations chosen to preserve the validity of the bounding box annotations: random horizontal and vertical flips, random 90 degree rotations, brightness and contrast jitter, and additive Gaussian noise.

\subsection{Model Pre Training}
\begin{figure}[htbp]
\centerline{\includegraphics[width=0.5\textwidth]{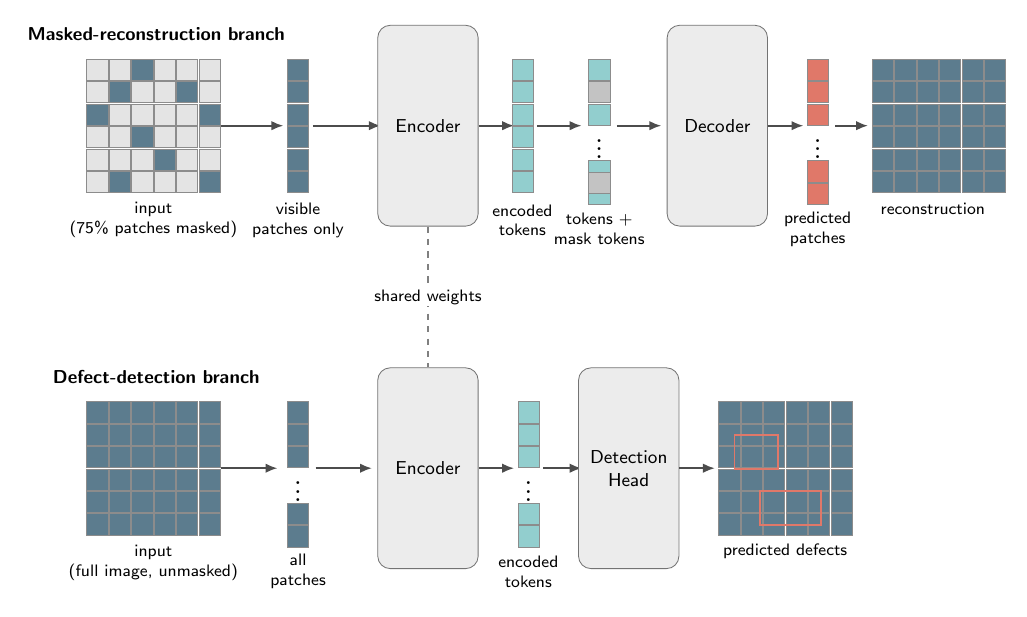}}
\caption{Model Architecture for Transformer based for regenerating input images and detecting the defects on the surface}
\label{fig:Model}
\end{figure}

The encoder is pretrained with a masked reconstruction objective, and an auxiliary detection branch sharing the same encoder is trained jointly. The two objectives are therefore evaluated on the same encoder with different passes within each training step and their losses are summed and optimized together. The overall architecture is shown in Figure~\ref{fig:Model}.

\subsubsection{Encoder and Decoder Architecture}

The encoder follows the Vision Transformer design. An input image of size
${224 \times 224}$ is divided into non overlapping
$16 \times 16$ patches, producing $196$ patches, each linearly projected
to a $768$ dimensional embedding by a convolutional patch embedding layer.
A learnable class token $x_{\text{cls}}$ is attached and learnable positional embeddings $E_{\text{pos}}$ are added, forming the input sequence.
\begin{equation}
z_0 = [\, x_{\text{cls}} \, ; \, x_p^{1} W_E \, ; \, x_p^{2} W_E \, ; \dots ; \, x_p^{N} W_E \,] + E_{\text{pos}},
\end{equation}
where $x_p^{i}$ is the $i$ th flattened patch and $W_E$ is the patch embedding
projection. The sequence is processed by a stack of six pre normalization
Transformer encoder layers with twelve attention heads.

\paragraph{Multihead self attention.}
Within each layer, the token representations are linearly projected into queries, keys, and values, and scaled dot product attention is computed between them, dividing by the square root of the per head dimension before the softmax to stabilize the scores. This is performed in parallel across twelve attention heads, whose outputs are concatenated and passed through a final output projection. Because every token attends to every other token, the encoder has a global receptive field, in contrast to the local receptive fields of convolutional networks.

\paragraph{Position wise feed forward network.}
Each attention sub layer is followed by a position wise feed forward network, a two layer multilayer perceptron with a GELU activation applied identically at every token position, combined with the attention sub layer through pre normalization and residual connections. A final layer normalization is applied after the last block, and the class token output is taken as the image level representation used for clustering.

\paragraph{Patch masking.}
During reconstruction pretraining, masking is performed as an index selection
operation rather than a learned transform. For each image, a random noise
vector is sampled and its ordering determines a random permutation of the $N$
patch indices; the first $N_{\text{keep}}$ are retained and the rest discarded,
where
\begin{equation}
N_{\text{keep}} = \lfloor (1 r)\, N \rfloor, \qquad r = 0.75,
\end{equation}
so that only $N_{\text{keep}} = 49$ of the $196$ patches are passed to the
encoder. Positional embeddings are added before this selection, so each
retained patch preserves its true spatial position, and the discarded
positions are later reintroduced in the decoder.

The decoder is deliberately lightweight and asymmetric with respect to the
encoder. It projects the encoder outputs to a $256$ dimensional
space and processes them with four pre normalization Transformer layers of
eight heads each, using the same attention and feed forward formulation given
above. Because only the visible patches are encoded and every masked position is represented by the same shared token in the decoder, the encoder processes far fewer tokens than the full image, which reduces pretraining cost.

\subsubsection{Optimization}

\begin{figure}[htbp]
\centerline{\includegraphics[width=0.5\textwidth]{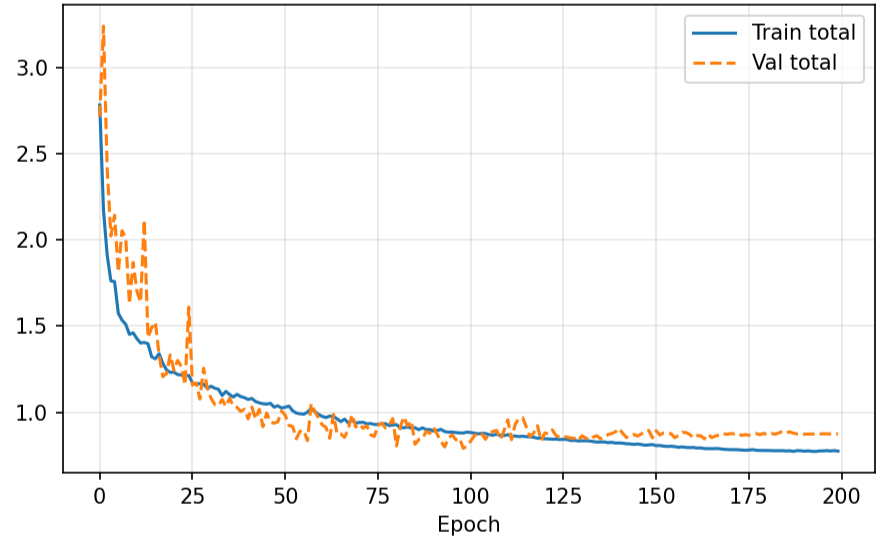}}
\caption{Pretraining Logs of the Masked Autoencoder to regenerate the input image}
\label{fig:training}
\end{figure}

The reconstruction and detection losses are summed and minimized in a single backward pass per batch. Training uses the AdamW optimizer with a cosine annealed learning rate schedule and gradient norm clipping, for 200 epochs at batch size 32. Training progress is monitored using the summation of loss together with the reconstruction MSE and MS SSIM, as shown in Figure~\ref{fig:training}.

\subsection{Dimensionality Reduction and Clustering}

After pretraining, the encoder is used as a fixed feature extractor. Each image
is passed through the encoder without masking, and its $768$ dimensional
class token embedding is taken as the image representation. The feature matrix
$X$, is standardized feature wise to zero mean and unit variance.

\subsubsection{Dimensionality Reduction}
The standardized features are projected to two dimensions using Uniform Manifold Approximation and Projection (UMAP) with a cosine distance metric, configured with 25 neighbors, a minimum distance of 0.1, and a fixed random seed of 742 for reproducibility. UMAP constructs a weighted nearest neighbor graph in the high dimensional space, in which the membership strength between two points decays with their cosine distance, and then optimizes a low dimensional embedding whose corresponding membership strengths match this structure by minimizing a fuzzy set cross entropy via stochastic gradient descent.

\subsubsection{Clustering}

The two dimensional embedding is partitioned into $K = 6$ groups, equal to the
known number of defect categories, using agglomerative hierarchical clustering
with Ward linkage. Starting from singleton clusters, the method iteratively
merges the pair whose union yields the smallest increase in total within cluster variance.

\subsubsection{Evaluation}

The resulting clusters are evaluated against the ground truth defect labels,
which are used only for scoring and play no role in the clustering. Denoting the ground truth label of sample $i$ by $y_i$, the predicted cluster by $c_i$, and the optimal cluster to label mapping by $\pi$, the clustering accuracy is
\begin{equation}
\text{ACC} = \frac{1}{n} \sum_{i=1}^{n} \mathbf{1}\!\left[\, y_i = \pi(c_i) \,\right],
\end{equation}
where $\mathbf{1}[\cdot]$ is the indicator function. Agreement is additionally
quantified using the Adjusted Rand Index (ARI)\cite{b15} and Normalized Mutual Information
(NMI)\cite{b16}. Given the contingency table between clusters and true classes, the ARI is
\begin{equation}
\text{ARI} = \frac{\sum_{ij} \binom{n_{ij}}{2} \Bigl[ \sum_i \binom{a_i}{2} \sum_j \binom{b_j}{2} \Bigr] / \binom{n}{2}}
{\tfrac{1}{2}\Bigl[ \sum_i \binom{a_i}{2} + \sum_j \binom{b_j}{2} \Bigr] \Bigl[ \sum_i \binom{a_i}{2} \sum_j \binom{b_j}{2} \Bigr] / \binom{n}{2}},
\end{equation}
where $n_{ij}$ is the number of samples in both cluster $i$ and class $j$, and
$a_i$, $b_j$ are the corresponding row and column sums. The NMI between the
cluster assignment $C$ and the true labeling $Y$ is
\begin{equation}
\text{NMI}(C, Y) = \frac{I(C; Y)}{\sqrt{H(C)\, H(Y)}},
\end{equation}
where $I(C; Y)$ is their mutual information and $H(\cdot)$ denotes entropy.

\section{Results and Discussions}

\begin{figure}[htbp]
\centerline{\includegraphics[width=0.55\textwidth]{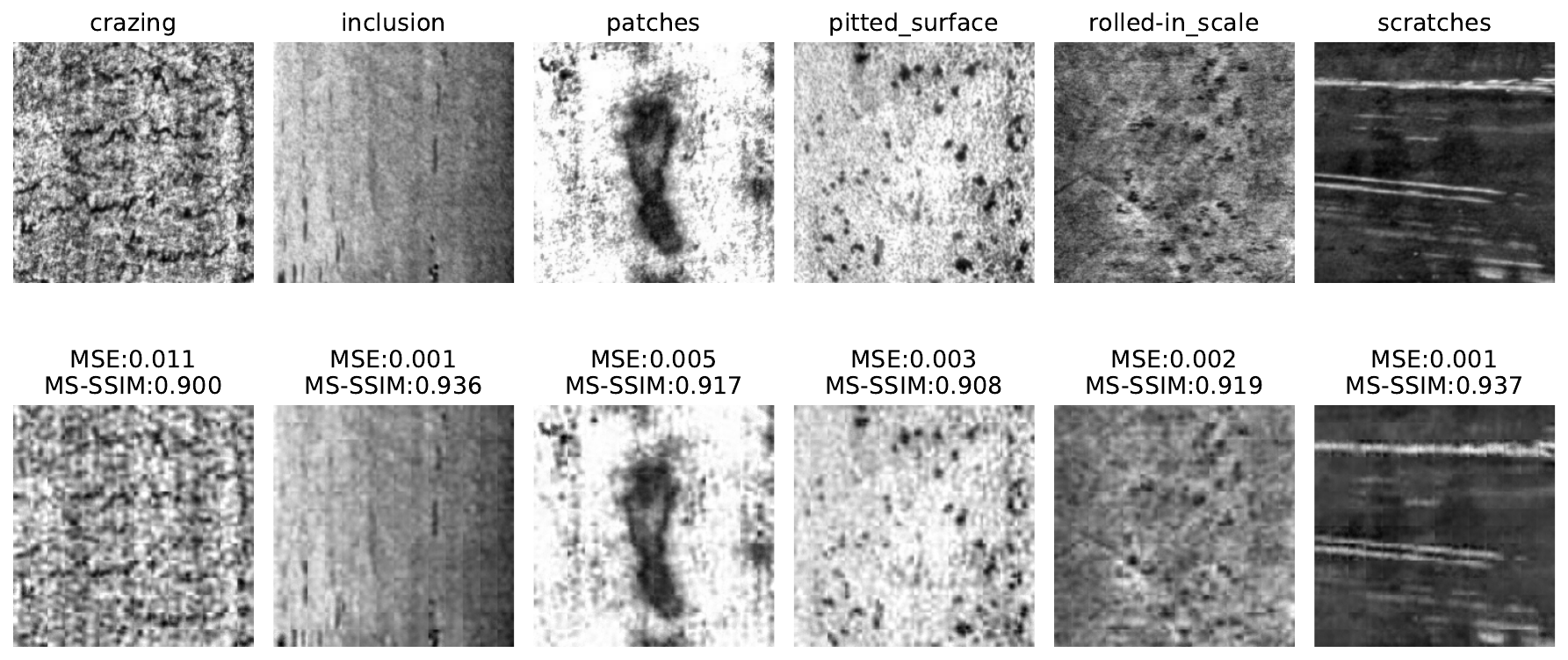}}
\caption{Decoder's ability to regenerate the input image from patches of encoder for each surface defect}
\label{fig:reconstruction}
\end{figure}

As shown in Figure~\ref{fig:reconstruction}, the model reconstructs the overall structure with a Multi-Scale Structural Similarity (MS-SSIM)\cite{b13} score of 0.92 and a mean squared error of 0.47; per image reconstruction errors for the examples as shown in Figure~\ref{fig:reconstruction} fall in the range of between 0.010 and 0.011. Pretraining therefore learned representations, adequate for the clustering step described next. Its effect is quantified directly in the comparison against a randomly initialized encoder below.

\begin{figure}[htbp]
\centerline{\includegraphics[width=0.4\textwidth]{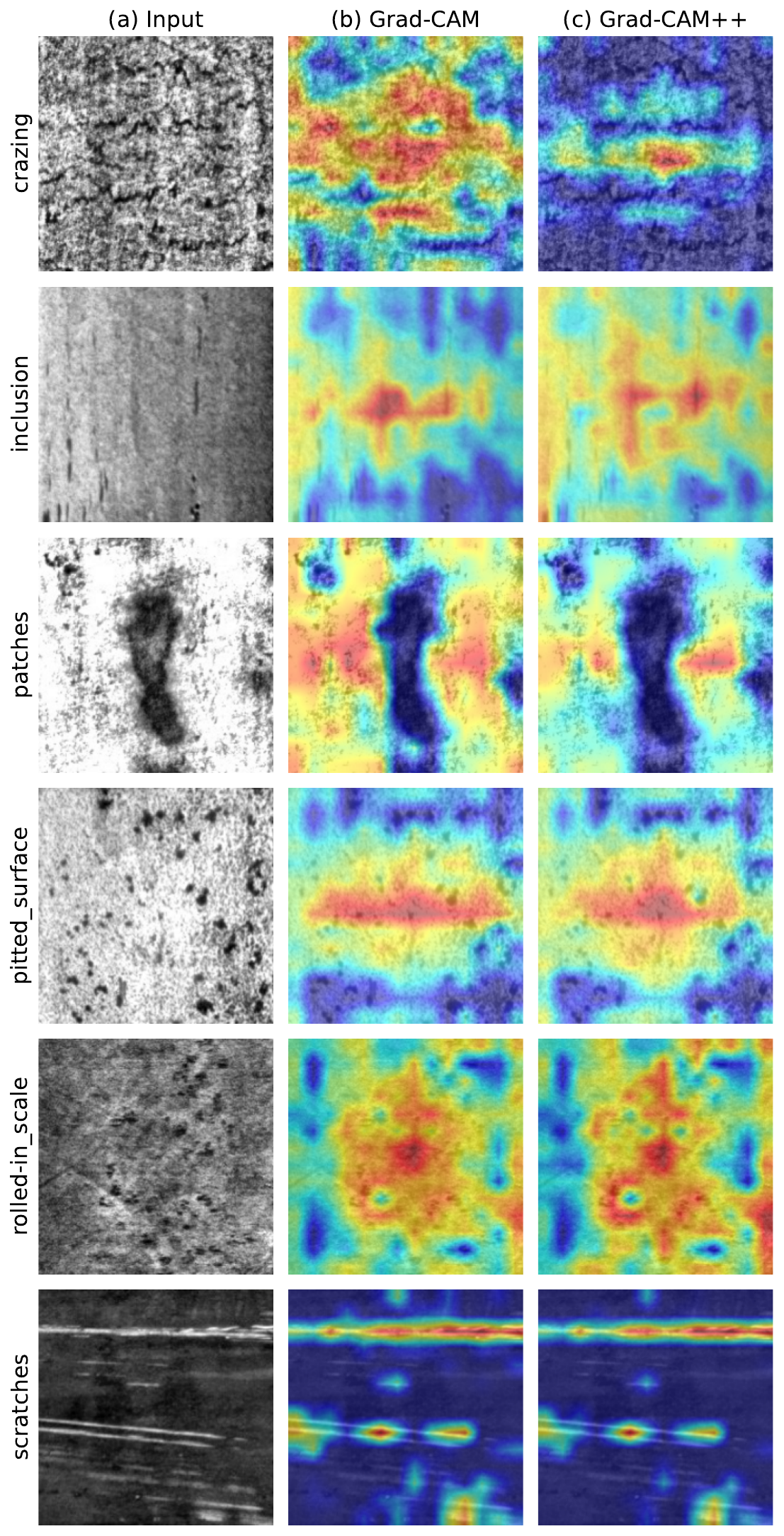}}
\caption{(a) Steel Surface Input passed to encoder (b) GradCAM and (c) GradCAM++ Results of the Encoder}
\label{fig:grad-cam}
\end{figure}

After Reconstruction, Encoder's attention was further examined using Grad CAM and Grad CAM++ without any classification head. As shown in Figure~\ref{fig:grad-cam}, both methods concentrate their attention on the defect region itself rather than on the surrounding, largely uniform steel surface: the pretrained encoder's representation is driven by defect relevant structure rather than by background texture. This pattern is consistent across defect categories, though the precise shape of the attention differs in ways that reflect the visual character of each defect. For patches, attention concentrates on the darker, irregularly shaped blob regions that define the defect, closely tracing their boundary. For scratches, attention sharply follows the bright, near linear streaks characteristic of this defect type, with little response elsewhere in the image. For inclusion, attention is drawn predominantly toward the straight, elongated defect markings running through the surface, again largely excluding the surrounding texture.

Comparing the two methods directly, Grad CAM and Grad CAM++ produce broadly similar attention patterns overall, but Grad CAM++ consistently yields a more spatially localized response on defects that are smaller or more spatially concentrated, reflecting its refinement over Grad CAM in weighting the contribution of individual activations rather than averaging them uniformly. This distinction is particularly visible for crazing, whose fine, web like cracks are distributed across the entire image rather than confined to a single region: under Grad CAM, attention is spread broadly across the image, tracking the diffuse extent of the defect, whereas under Grad CAM++, attention is noticeably more concentrated toward the central region of the image. The learned representation therefore attends to defect relevant structure across a range of defect geometries, from the sharply localized patches and scratches to the more spatially distributed texture of crazing.

The pretrained encoder's class token features were standardized and reduced to two dimensions using UMAP with a cosine distance metrics. The two dimensional embedding was then partitioned into six groups, matching the known number of defect categories, using agglomerative clustering with Ward linkage. Because cluster indices carry no inherent correspondence to the true defect categories, each cluster was matched to a ground truth class by solving an optimal one to one assignment with the Hungarian algorithm\cite{b14}; under this matching, the clustering achieved an accuracy of 91.33\%. An Adjusted Rand Index (ARI) of 0.815 and a Normalized Mutual Information (NMI) of 0.834, both computed independently of the Hungarian assignment and both accounting for chance agreement, point to the same conclusion: the correspondence between clusters and true categories is far stronger than a random grouping of the same size would produce. The clustering also attained a silhouette score of 0.726 on the UMAP embedding, an internal measure computed without reference to the true labels; the discovered groups are compact within themselves and clearly separated from one another. 

\begin{figure}[htbp]
\centerline{\includegraphics[width=0.5\textwidth]{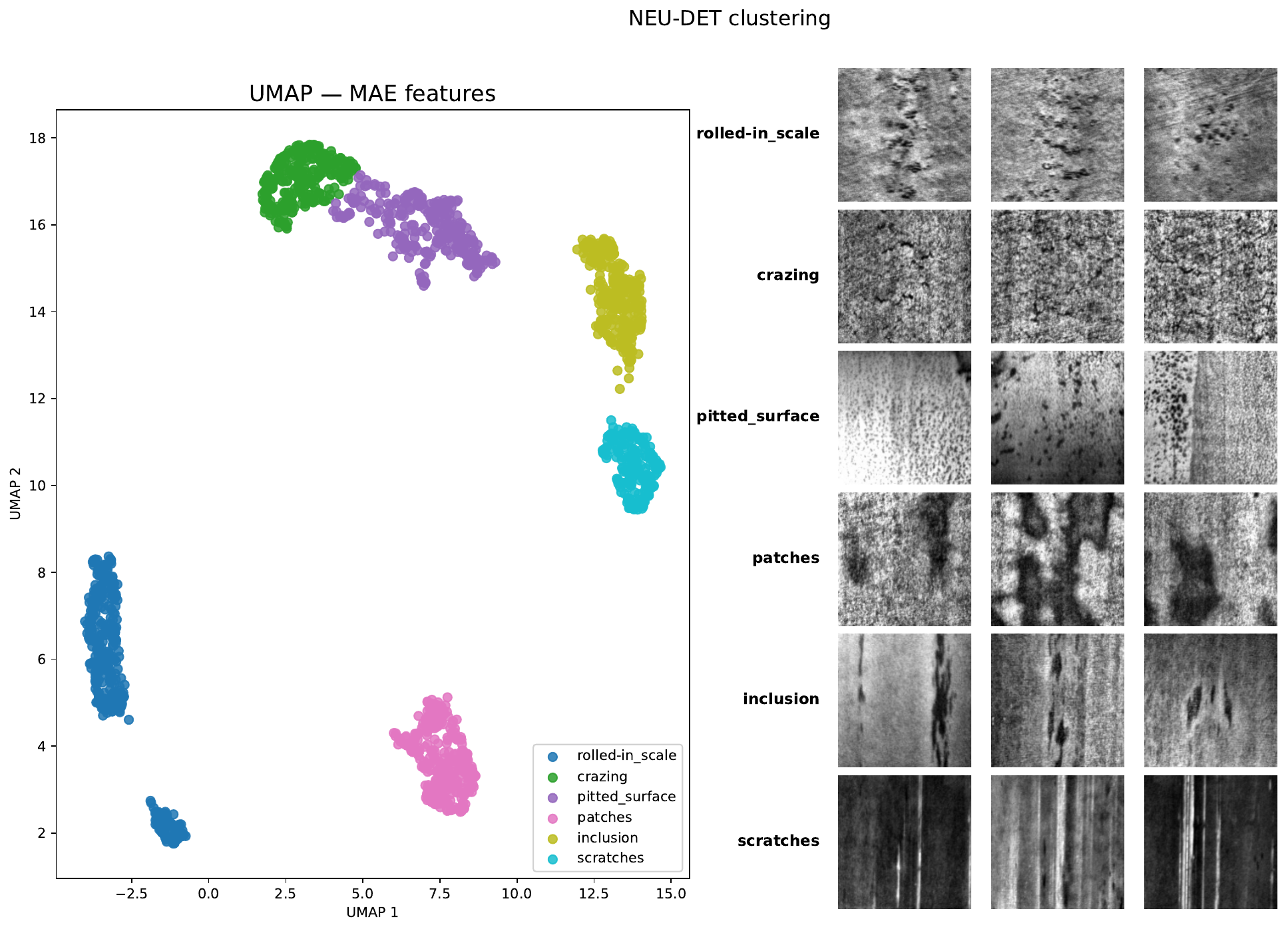}}
\caption{Clusters created by Agglomerative clustering algorithm and sample images present in each cluster}
\label{fig:clustering}
\end{figure}

Figure~\ref{fig:clustering} shows the resulting two dimensional embedding, in which the six defect categories form clearly separated groups, alongside representative images drawn from each cluster. Moreover, it can also be seen from the figure that clusters of each defect are highly seperated from one another. The representative samples drawn from each cluster and displayed alongside the embedding further illustrate this separation: images grouped within the same cluster share a consistent and visually recognizable defect appearance, while images from different clusters correspond to visibly distinct defect types. This visual coherence within clusters, together with the clear spatial separation between them, indicates that the pretrained encoder produces feature representations in which the six defect categories are largely distinguishable on the basis of the learned features alone.

\begin{figure}[htbp]
\centerline{\includegraphics[width=0.4\textwidth]{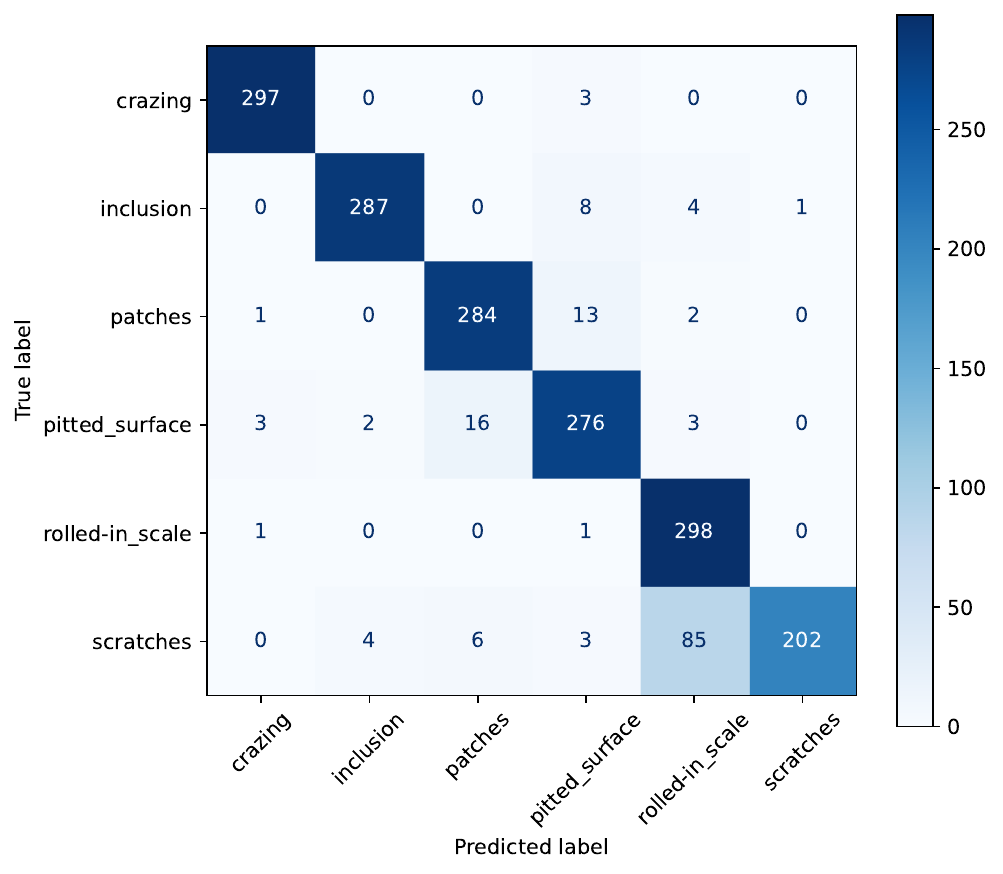}}
\caption{Confusion Matrix for true vs predicted labels}
\label{fig:conf-mat}
\end{figure}

The confusion matrix between true labels and Hungarian matched clusters, shown in Figure~\ref{fig:conf-mat}, provides a finer grained view of this result. It provides insights about the error rate and precision of the encoder where it rigorously classifies the defects from the image correctly. Five of these six categories are recovered with high purity: crazing, inclusion, patches, pitted surface, and rolled in scale each have the large majority of their samples between 276 and 298 of roughly 300 correctly assigned to a single cluster. The principal source of error is the scratches category, of which 85 of 300 samples are absorbed into the rolled in scale cluster because they share a linear, directional surface texture. A smaller degree of confusion is also observed between pitted surface and patches (16 samples). These two confusions account for the majority of the gap between the observed 91.3\% accuracy and perfect agreement. Overall, the learned representations are able to precisely distinguish between these surface defects.

The identical UMAP and Agglomerative clustering procedure was applied in order to analyze the impact of pretraining and data augmentation, to two additional feature representations of the same dataset with raw pixel intensities and features from an encoder of identical architecture with randomly initialized weights. The second baseline is particularly informative, as it isolates whether the Transformer architecture alone imposes useful structure on its output.
As summarized in Table~\ref{tab:baselines}, both baselines perform substantially worse than the pretrained encoder: raw pixels reach only 34.7\% accuracy and the randomly initialized encoder 40.8\%, compared with 91.3\% for the pretrained encoder, with silhouette scores of 0.432, 0.392, and 0.726 respectively. The randomly initialized encoder performs only marginally better than raw pixels despite sharing the full architecture; architecture alone therefore confers little of the observed clustering quality. Because the dimensionality reduction and clustering algorithms are held fixed identically across all three conditions, this large and consistent gap demonstrates that the clustering performance is driven specifically by the representation learned during masked autoencoder pretraining, and not by the downstream pipeline itself.
\begin{table}[t]
\centering
\caption{Clustering performance of the MAE pretrained encoder compared with baselines, using the identical UMAP + Agglomerative clustering pipeline.}
\label{tab:baselines}
\begin{tabular}{lcc}
\hline
\textbf{Representation} & \textbf{Accuracy} & \textbf{Silhouette} \\
\hline
Raw pixels & 0.3472 & 0.4318 \\
Random init encoder & 0.4078 & 0.3915 \\
MAE pretrained encoder & \textbf{0.9133} & \textbf{0.7259} \\
\hline
\end{tabular}
\end{table}

\section{Conclusion and Future Works}
A Transformer based Masked Autoencoder for learning steel surface defect representations was jointly pretrained with an auxiliary detection objective. The encoder reconstructed validation images accurately, achieving a validation MS-SSIM of approximately 0.92 and mean squared error of approximately 0.47, with Grad CAM and Grad CAM++ confirming attention on defect regions rather than background. Clustering the pretrained features with UMAP and Agglomerative clustering achieved 91.33\% accuracy, an ARI of 0.815, and an NMI of 0.834, outperforming raw pixel and randomly initialized baselines by a wide margin: the gain comes from pretraining rather than from the clustering pipeline. Scratches, often confused with rolled in scale, was the main source of residual error. The detection objective served only to shape the encoder and was not evaluated as a detector; improving it correcting the target encoding, adopting an IoU based loss, and using a finer detection grid is a natural next step, both as a detector in its own right and as a possible route to stronger clustering, since both objectives shape the same encoder.


\begin{thebibliography}{00}

\bibitem{b1} K. Song and Y. Yan, ``A noise robust method based on completed local binary patterns for hot rolled steel strip surface defects,'' \textit{Applied Surface Science}, vol. 285, pp. 858--864, 2013.
\bibitem{b2} X. Lv, F. Duan, J. Jiang, X. Fu, and L. Gan, ``Deep metallic surface defect detection: The new benchmark and detection network,'' \textit{Sensors}, vol. 20, no. 6, p. 1562, 2020.
\bibitem{b3} S. Zhou, Y. Cai, Z. Zhang, and J. Yin, ``MESC DETR: An improved RT DETR algorithm for steel surface defect detection,'' \textit{Electronics}, vol. 14, no. 11, p. 2232, 2025.
\bibitem{b4} K. He, X. Chen, S. Xie, Y. Li, P. Doll\'ar, and R. Girshick, ``Masked autoencoders are scalable vision learners,'' in \textit{Proc. IEEE/CVF Conference on Computer Vision and Pattern Recognition (CVPR)}, 2022, pp. 16000--16009.
\bibitem{b5} A. Dosovitskiy \textit{et al.}, ``An image is worth 16x16 words: Transformers for image recognition at scale,'' in \textit{Proc. International Conference on Learning Representations (ICLR)}, 2021.
\bibitem{b6} L. McInnes, J. Healy, and J. Melville, ``UMAP: Uniform manifold approximation and projection for dimension reduction,'' \textit{arXiv preprint arXiv:1802.03426}, 2018.
\bibitem{b7} R. R. Selvaraju, M. Cogswell, A. Das, R. Vedantam, D. Parikh, and D. Batra, ``Grad CAM: Visual explanations from deep networks via gradient based localization,'' in \textit{Proc. IEEE International Conference on Computer Vision (ICCV)}, Venice, Italy, 2017, pp. 618--626.
\bibitem{b8} A. Chattopadhay, A. Sarkar, P. Howlader, and V. N. Balasubramanian, ``Grad CAM++: Generalized gradient based visual explanations for deep convolutional networks,'' in \textit{Proc. IEEE Winter Conference on Applications of Computer Vision (WACV)}, Lake Tahoe, NV, USA, 2018, pp. 839--847.
\bibitem{b9} W. Zhu, H. Zhang, C. Zhang, X. Zhu, Z. Guan, and J. Jia, ``Surface defect detection and classification of steel using an efficient Swin Transformer,'' \textit{Advanced Engineering Informatics}, vol. 57, p. 102061, 2023.
\bibitem{b10} R. Xu, R. Hao, and B. Huang, ``Efficient surface defect detection using self supervised learning strategy and segmentation network,'' \textit{Advanced Engineering Informatics}, vol. 52, p. 101566, 2022.
\bibitem{b11} Y. He, K. Song, Q. Meng, and Y. Yan, ``An end to end steel surface defect detection approach via fusing multiple hierarchical features,'' \textit{IEEE Transactions on Instrumentation and Measurement}, vol. 69, no. 4, pp. 1493--1504, 2019.
\bibitem{b12} Y. Gao, G. Lv, D. Xiao, X. Han, T. Sun, and Z. Li, ``Research on steel surface defect classification method based on deep learning,'' \textit{Scientific Reports}, vol. 14, p. 8254, 2024.
\bibitem{b13} Z. Wang, E. P. Simoncelli, and A. C. Bovik, ``Multiscale structural similarity for image quality assessment,'' in \textit{Proc. 37th Asilomar Conference on Signals, Systems and Computers}, Pacific Grove, CA, USA, 2003, pp. 1398--1402.
\bibitem{b14} H. W. Kuhn, ``The Hungarian method for the assignment problem,'' \textit{Naval Research Logistics Quarterly}, vol. 2, no. 1--2, pp. 83--97, 1955.
\bibitem{b15} L. Hubert and P. Arabie, ``Comparing partitions,'' \textit{Journal of Classification}, vol. 2, no. 1, pp. 193--218, 1985.
\bibitem{b16} N. X. Vinh, J. Epps, and J. Bailey, ``Information theoretic measures for clusterings comparison: Variants, properties, normalization and correction for chance,'' \textit{Journal of Machine Learning Research}, vol. 11, pp. 2837--2854, 2010.
\bibitem{b17} George, Ajo Babu, and Sadhvik Bathini. "Grad-cam \& grad-cam++ for explainable oral squamous cell carcinoma detection using deep learning on orthopantomograms." \textit{2025 International Conference on Sensors and Related Networks (SENNET) Special Focus on Digital Healthcare} (64220). IEEE, 2025.
\bibitem{b18} Govind, A., Chandra P. Dubey, and Jyotiranjan Ray. "Multimodal Deep Learning Based Prediction of Volumetric Deformation of Barren Island using InSAR Inferrograms." \textit{Authorea Preprints} (2025).
\end{thebibliography}
\end{document}